# Zero Cost Improvements for General Object Detection Network

Shaohua Wang[1], Yaping Dai[2]

1. School of Automation, Beijing Institute of Technology, Beijing 100081, China
E-mail: wang.shaohua@bit.edu.cn

2. School of Automation, Beijing Institute of Technology, Beijing 100081, China
E-mail: daiyaping@bit.edu.cn

**Abstract:** Modern object detection networks pursuit higher precision on general object detection datasets, at the same time the computation burden is also increasing along with the improvement of precision. Nevertheless, the inference time and precision are both critical to object detection system which needs to be real-time. It is necessary to research precision improvement without extra computation cost. In this work, two modules are proposed to improve detection precision with zero cost, which are focus on FPN and detection head improvement for general object detection networks. We employ the scale attention mechanism to efficiently fuse multi-level feature maps with less parameters, which is called SA-FPN module. Considering the correlation of classification head and regression head, we use sequential head to take the place of widely-used parallel head, which is called Seq-HEAD module. To evaluate the effectiveness, we apply the two modules to some modern state-of-art object detection networks, including anchor-based and anchor-free. Experiment results on coco dataset show that the networks with the two modules can surpass original networks by 1.1 AP and 0.8 AP with zero cost for anchor-based and anchor-free networks, respectively. Code will be available at https://git.io/JTFGl.

## 1 INTRODUCTION

Object detection is a fundamental task in computer vision, which is beneficial to downstream tasks such as instance segmentation, object tracking, face recognition, and so on. With the development of deep learning and CNN (Convolutional Neural Network) [1,2,3], the object detection networks based on CNN have been the mainstream methods in object detection domain. The object detection networks can be divided into two branches according to the detection process. The two-stage methods [4,5,6,7] have high detection precision but compromise on detection speed. They model objection detection task as two-stage procedure: obtaining region proposals first, then fining tuning proposals to generate bounding box and predict the category. On the contrary, the one-stage methods [8,9,10] mainly focus on enhance the detection speed of networks. Recently, the state-of-art one-stage networks have achieved comparable precision as two-stage networks, while with higher detection speed. Since in practical application scenarios, the real-time performance of the object detection system is highly required. In this paper, we mainly study improving the precision of one-stage networks while maintaining their inference speed.

Most of the recent improvements on precision of one-stage networks have come at the expense of increasing inference time, which go against the original design intention of one-stage networks. PANet [11] adds an additional bottom-up path behind top-down path of original FPN [12], and the additional convolution layers bring more parameters and slow down the inference speed. FSAF [13] fuses anchor-free branch in anchor-based networks and more parameters are also introduced. ASFF [14] proposes adaptive spatial feature fusion mechanism, which is applied on feature maps of FPN's output. This module increases the AP (average precision) of YOLOv3 [10] by 6.3 but reduces the FPS by 5.4. We proposed two modules different from these methods that can be applied on general objection detection networks to improve detection precision with zero extra computational cost.

We propose SA-FPN (Scale Attention FPN) module to enhance multi scale feature fusion in FPN and Seq-Head (Sequential Head) module to efficiently utilize the correlation between classification and regression head.

The original FPN uses top-down architecture and lateral connections to adjacent level feature maps to detect objects with various scales. Since its only fuse adjacent feature map, the ability of the network to fuse features with large scale differences is limited. Such as fusing semantic information of highest scale feature map to lowest scale feature map which has high resolution. Recently, some excellent works have proved the effectiveness of multi-scale feature fusion technology in object detection domain. Such as DetectoRS[15], which has taken over the MS-COCO[16] leaderboard, proposes to replace convolution with SAC (Switchable Atrous Convolution). The key idea of SAC is that using different atrous rate to generate different scale feature maps and then fusing them adaptively by attention mechanism. Since the effectiveness of the multi-scale feature fusion has been proved, we introduce its fusion conception into FPN structure. Our proposed SA-FPN module takes all scale feature maps as inputs, adaptively fusing whole feature maps to generate each newly feature maps and the weights are learned from input feature maps individually. The overall architecture of SA-FPN module is shown in Figure 1, which looks like a full connected layer but actually quite different. Each input of SA-FPN module is a feature map rather than a single variable and the fusion procedure is more complicated as shown in Figure 2. The fusion procedure is inspired from SKNet[20] and to better embed it into FPN we modified some parts. Compared to original FPN structure, SA-FPN module can efficiently fuse all scale feature maps with higher precision and less parameters. As shown in Table 1, SA-FPN module improves baseline's average precision by 0.8 points but reduces the number of parameters by 1.2 million. The reduced parameters mainly come from 1x1 convolution after adjacent fusion in original FPN which is used to refine simple fused feature maps. In SA-FPN module, the newly feature maps have been well fused thanks to the scale attention mechanism. Therefore, there is no need to add additional convolutional layer after fusion. The ablation study in Table 1 also proves that the removal of 1x1 convolution can bring higher precision.

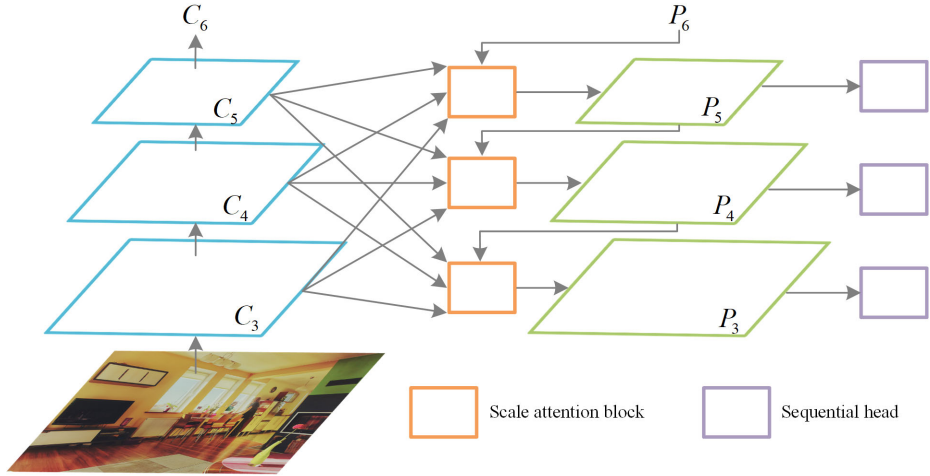

Fig 1. The overall architecture of one-stage network equipped with SA-FPN module and sequential head module. To simplify, only the first three low scale levels of total 7 levels are shown. The core of SA-FPN module is scale attention block which is represented as orange box. These blocks take feature maps of all scale levels as inputs and generate corresponding new feature map, the weights are independent for each block. The sequential head module is represented as purple box, and its weights are shared across all scale level.

The detection head of general object detection networks is parallel head, which is widely used but has obvious disadvantages. The parallel head ignores the correlation between classification and regression of bounding box. For instance, the height of a person's bounding box should be larger than its width, and the width of car's bounding box should be larger than its height. In short, the classification results can provide prior information for regression task. To efficiently enhance the correlation between category and bounding box of object, we propose a sequential head which is called Seq-HEAD module. According to the above analysis, we put classification head first and the regression head followed. Such that the feature maps learned from classification head can be treated as prior information of regression head. The feature maps are efficiently reused, and the correlation between size prediction and category prediction has been enhanced. As shown in Table 4, the sequential head brings 0.3 points gain of average precision for anchor-based network RetinaNet[22] and 0.6 points gain for anchor-free network FCOS[23], while no additional parameters are introduced, since we retain the number of layers in original parallel head.

In this paper, two zero cost modules for general object detection networks are proposed. Our modules are suitable for both anchor-based and anchor-free networks, with AP improvement of 1.1 and 0.8 points respectively. Noticing that the gain of anchor-free networks is not as large as that of anchor-based networks, we will detail reasons in ablation study.

Finally, our contributions can be summarized as follows:

- We propose SA-FPN module, a lighter but more efficient alternative to original FPN structure. It uses scale attention mechanism to adaptively fuse all scale feature maps from backbone, and generate newly feature map dependently.
- We propose Seq-Head module to make full use of the correlation between regression and classification head, increasing precision while not introducing extra parameters.
- We conduct experiments on both anchor-based and anchor-free networks, the precision improvements are observed on all of them and the robustness of two modules is demonstrated.

## 2   RELATED WORKS

**FPN structure.** PANet[11] adds an additional bottom-up path after top-down path of original FPN, which shortens the information path between lower layers and topmost feature, further enhances the localization capability of the entire feature hierarchy. NAS-FPN[17] uses Neural Architecture Search technology to discover a new feature pyramid architecture in search space which covers all cross-scale connections. The finally searched NAS-FPN also contains bottom-up and top-down path but with more complex connections. EfficientDet[18] redesigns the bottom-up path in PANet by removing unnecessary connections, such as nodes that only have one input edge, and adding extra edges from the original input to output nodes when they are at same level. These excellent works have enhanced the ability of feature fusion of original FPN and improved the precision, but introduced more parameters. The additional parameters slow down the inference speed of networks, and obey the real-time requirement of object detection system. In this paper, we apply scale attention mechanism on original FPN, making it more accurate with fewer parameters.

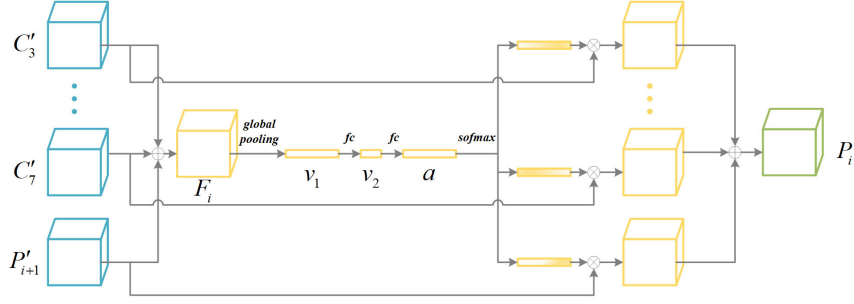

Fig 2. The structure of scale attention block of SA-FPN. Some input and middle feature maps are omitted and replaced by dots for simplicity.

**Attention Mechanism.** SENet[19] proposes channel attention mechanism for feature maps after convolution operation, which adaptively recalibrates channel-wise feature responses by explicitly modelling interdependencies between channels. SKNet[20] uses multi-branch channel attention to fuse multiple scales information generated by applying different kernel size to one input feature map, called SKConv. Different attentions on these branches yield different sizes of the effective receptive fields of neurons in the fusion layer. ASFF[14] proposes adaptively spatial feature fusion with learned attention weight to deal with inconsistency across different feature scales after FPN. ResNeSt[21] presents split-attention block in original ResNet that enables attention across feature-map groups. Its split-attention module is a kind of variant of SKConv. DETR [22] introduces attention of natural language processing into objection detection. DetectoRS[15] proposes Switchable Atrous Convlution to replace traditional convolution, it uses different atrous rates to generate multi-scale feature maps, and fuses them with learned attention weight.

## 3 ZERO COST IMPROVMENTS

We first provide an overview of the approach. The overall network architecture with proposed modules is shown in Figure 1. The network is based on RetinaNet[23], in which we apply SA-FPN module to replace the origin FPN[12] structure, and modify the parallel heads to the proposed Seq-HEAD module. We first generate 5 scales feature maps $\{C_3, C_4, C_5, C_6, C_7\}$ with stride of $\{8, 16, 32, 64, 128\}$ from backbone such as ResNet[3]. The feature maps $\{C_3, C_4, C_5\}$ are extracted from the last three stage's output of Resnet, $\{C_6, C_7\}$ are generated by adding 3x3 convolution layer on $C_5$ and $C_6$, respectively. The proposed two modules both have less or equal parameters but with higher precision comparing to their counterparts. The two modules are plug and play, which means they can be used on general objection networks that contain FPN structure and detection head to obtain zero cost AP improvement.

### 3.1 SA-FPN module

Original FPN module uses top-down structure and lateral connection to enhance feature map, which only fuses the information of adjacent scale. Considering each scale feature map may contain useful information for other scale feature map, our SA-FPN module uses scale attention mechanism to enhance each feature map by integrating information from all scales feature maps.

As shown in Figure 2, the feature maps generated by backbone are inputs. The scale attention module learns attention weights from input feature maps $\{C_3, C_4, C_5, C_6, C_7\}$ and integrates them using learned weights to generate newly feature maps $\{P_3, P_4, P_5, P_6, P_7\}$. The whole procedure in scale attention module can be formulated as follows:

$$P_i = \begin{cases} SA(C'_3 \sim C'_7) & i = 7 \\ SA(C'_3 \sim C'_7, P'_{i+1}) & i = 3, 4, 5, 6 \end{cases} \quad (1)$$

where $i$ is the scale level of feature map and $SA()$ is the scale attention block. The $C'_3$ represents $C_3$ is resized to the shape of target new feature map $P_i$, as well as the $C'_7$ and $P'_{i+1}$. Since the newly feature maps of FPN are generated in a top-down manner, for level 7 the scale attention block only takes $C'_3 \sim C'_7$ as inputs, and other levels take $C'_3 \sim C'_7$ and higher level newly generated feature map $P'_{i+1}$ as inputs.

To use scale attention block generating newly feature map, the inputs need to have the same shape with target feature map. Instead of using up-sample or down-sample convolution, we use bilinear interpolation for all resized operation in order not to introduce additional parameters which goes against our zero cost design. As shown in Figure 2, all resized input feature maps are first added element-by- element to form an aggregated feature map $F_i \in \mathbb{R}^{H_i \times W_i \times C}$:

$$F_i = \sum_{j=3}^{7} C'_j + P'_{i+1} \qquad (2)$$

then we use global average pooling at channel dimension to obtain channel-wise global information vector $v_1 \in \mathbb{R}^C$. The $C$ is channel dimension of feature map, which is fixed as 256 in all of our experiments. To reduce the number of parameters, a fc layer is following pooling layer to reduce the dimension of the global information vector to $v_2 \in \mathbb{R}^d$. Then another fc layer is applied to construct scale attention vector $a \in \mathbb{R}^C$. A softmax layer is used to generate the final attention weight for each input feature map. All input feature maps are combined with attention weights to generate the target fused feature map $P_i$.

Our motivation comes from SKConv (selective kernel convolution) of SKNet[20], we state the differences between our module and SKConv. First, the SKConv uses convolution with different kernel sizes to obtain different scale feature maps as inputs. Our module takes the advantage of multi-stage structure of backbone, naturally taking the output of each stage as inputs. Second, our module combines 5 or 6 scales feature maps while SKConv only combining 2. Third, the SKConv is employed in backbone to enable neurons to adaptively adjust their receptive field size. Our module is employed in FPN structure to enhance the feature fusion.

### 3.2 Sequential Head module

The detection head of general one-stage object networks consist of classification head and regression head, as shown in Figure 3(a). The two heads are parallel in forward inference process, which means the model predicts category and bounding box of object independently. However, the category and bounding box are naturally associated so predicting them separately misses this correlation, which may be the main reason of low precision of previous networks. For instance, the bounding box of cars should be width longer than height, which is opposite to person. Our idea is that the category information of object can provide prior knowledge to regress its bounding box.

We propose the sequential head which considers the relation between the classification and regression head so that the category information can be efficiently used as prior to enhance the precision of bounding box regression. As shown in Figure 3(b), the feature map outputted by SA-FPN is firstly fed into the feature extraction tower of classification head, which contains 4 convolution layers and is represented by blue box. The yellow layer at first row is the last layer of classification head which is responsible for predicting the probability of the object's existence at each location of feature map. We regard the feature map before last output layer in classification head as the prior category feature map for regression head, which is fed into regression head as its input. The provided prior information elevates the ability of regression head, which is proved in Section 4.2.

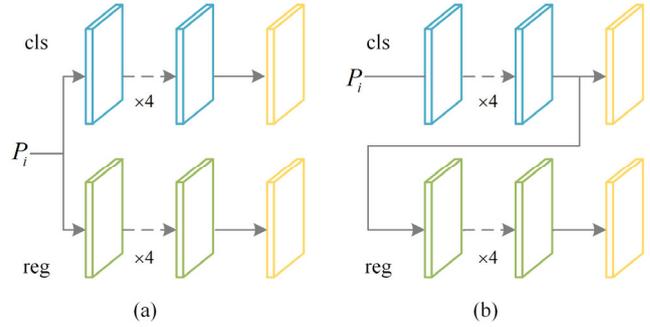

Fig 3. The difference of original parallel heads and our sequential heads. (a)The structure of parallel heads, in which classification and regression head are independent and parallel. (b)The structure of sequential heads, in which the regression head is behind of classification head. The module is shared across all scales output feature maps of SA-FPN.

## 4 EXPERIMENTS

The experiments are performed on the MS-COCO 2017 datasets[15]. We use the *trainval35k* split to train the model that contains 80k train images and 35k subset of images from the 40k image *val* split, and then evaluate model on the *minival* split which contains remaining 5k images from *val* split. We first describe the implementation details about training hyperparameter configuration in Section 4.1. Then the ablation studies in Section 4.2 demonstrates the effectiveness of proposed two modules. Finally, we employ the two proposed modules on both anchor-based and anchor-free networks, and discuss the AP gain without extra computational cost.

### 4.1 Implementation Details

**Initialization:** We use the ResNet-50 and ResNet-101 as model's backbones. If not otherwise specified, the default backbone in experiments is ResNet-50. The two backbones are all pre-trained on ImageNet1k[25], we use the pre-trained model provided by PyTorch[26]. The classification head and regression head are initialized by the procedure in RetinaNet[23]. The bias of fc layers of scale attention module is initialized to zero and the initial weight is normalized to have Gaussian distribution with mean of zero and standard deviation of 0.01.

**Optimization:** We employ RetinaNet and FCOS[24] as our anchor-based and anchor-free baseline, respectively. The training scheme of RetinaNet trains the model for total 90k iterations with a minibatch size 16. The learning rate is initialized to 0.01 and divided by 10 at 60k and 80k iterations. They use 8 GPUs to train the model (2 images per GPU). But since the limitation of hardware resource, we have to decrease the minibatch size and increase the total iterations to achieve comparable effect, which is demonstrated to be effective in Detectron[25]. We train the model for total 180k iterations with a minibatch size 8 (4 images per GPU). The learning rate is initialized to 0.005 and divided by 10 at 120k and 160k iterations. We use SGD optimizer, weight decay of 0.0001 and momentum of 0.9 are used. In both training and testing procedure, the short side of input image is resized to 800, and the longer side is resized less than 1333.

**Evaluation:** For the evaluation, we report the standard COCO evaluation metric of Average Precision (AP) as well as $AP_{50}$ and $AP_{75}$. We also report COCO-style $AP_S$, $AP_M$ and $AP_L$ on objects of small (less than 32x32), medium (from 32x32 to 96x96), and large (greater than 96x96) sizes. Besides, to demonstrate the zero cost design of our two model, we count the number of parameters of models which are equipped with the two proposed modules.

### 4.2 Ablation Studies

The ablations studies are all conducted on RetinaNet with ResNet-50 as backbone. And the models are evaluated on COCO 2017 *val* split.

**SA-FPN.** Ablation studies are conducted on SA-FPN to find out the best combination of inputs in scale attention blocks. We experiment three combination schemes, the details are explained as follows:

- **ALS** (All level scheme). In this scheme, the feature maps of all scales are inputs, and the 3x3 conv layers after newly generated feature map from original FPN are retained. Since this scheme is equal to add additional structure on original FPN, the parameters also increase.
- **ALS-Light** (Light version of all level scheme). The light version pursues less parameters than original FPN, which discard the last 3x3 conv layer in all scale of FPN's output.
- **LLS** (Low level scheme). The high level feature maps {C6, C7} are generated by means of adding single additional conv layer rather than being extracted from stage's output of backbone. For these two feature maps, we only replace the lateral connection in FPN with scale attention block, and keep the element-wise sum operation with high level newly generated feature map.

The results of ablation study about different combination schemes are shown in Table 1. We can see that all of the three schemes obtain AP improvement compared to baseline, with the improvements of 0.7, 0.9 and 0.3, respectively. These results demonstrate the effectiveness of our improved schemes about FPN. In addition to this, we also observe an interesting phenomenon: the ALS-Light scheme has the least parameters but the best performance, which surpasses baseline by 0.9 AP with 1.2M short of parameters. On the contrary, the ALS and LLS scheme have more parameters than baseline but only little AP improvement. This phenomenon indicates that the AP improvement of the ALS-Light scheme does not depend on the increase of the number of parameters, but is the result of the dynamic integration of scale attention mechanism in SA-FPN. The ALS scheme promotes the $AP_S$ by 1.9 points and the $AP_L$ by 0.9 points, which confirms that the SA-FPN module fuses multi-scale feature maps efficiently so that the model's ability to detect objects with various sizes has improved completely. But its parameters surpass the baseline which disobeys our zero cost design, and its AP is 0.2 points behind the ALS-Light scheme. Hence, we select

Table1. Comparisons of different combination schemes of inputs of scale attention block in SA-FPN. (**Params:** the number of parameters of models)

| Scheme | AP | $AP_{50}$ | $AP_{75}$ | $AP_S$ | $AP_M$ | $AP_L$ | Params /M |
|---|---|---|---|---|---|---|---|
| Baseline | 36.1 | 55.1 | 38.4 | 19.4 | 39.6 | 48.8 | 37.7 |
| ALS | 36.8 | 56.4 | 39.2 | **21.7** | 40.3 | **49.7** | 39.4 |
| ALS-Light | **37.0** | 56.4 | **39.7** | 20.3 | **40.9** | 49.4 | **36.5** |
| LLS | 36.4 | 55.5 | 39.0 | 20.6 | 40.0 | 48.3 | 37.9 |

Table2. Comparisons of different order schemes of classification head and regression head in sequential heads module. (**Params:** the number of parameters of models)

| Scheme | AP | $AP_{50}$ | $AP_{75}$ | $AP_S$ | $AP_M$ | $AP_L$ | Params /M |
|---|---|---|---|---|---|---|---|
| Baseline | 36.1 | 55.1 | 38.4 | 19.4 | 39.6 | 48.8 | 37.7 |
| Cls-First | **36.4** | **55.6** | **39.1** | **19.8** | **39.6** | **48.2** | 37.7 |
| Reg-First | 35.8 | 54.2 | 38.2 | 18.9 | 39.1 | 47.4 | 37.7 |

Table3. Real-time comparisons between the two Seq-HEAD schemes and the baseline. (**FPS**: Frames Per Second.)

| Scheme | Baseline | Cls-First | Reg-First |
|---|---|---|---|
| FPS | 15.8 | 15.8 | 15.8 |

the ALS-Light scheme as the default scheme of scale attention block in SA-FPN.

**Seq-HEAD.** The ablation studies of sequential head module are conducted to determine whether the classification head should be in front of or behind the regression head. We experiment two schemes about the order of two heads: **Cls-First** (classification head first scheme) and **Reg-First** (regression head first scheme). The experimental results are shown in Table 2. Since the sequential heads only changes the combination order of the layers and no additional parameters are introduced, the models equipped with two schemes both have equal number of parameters compared with baseline. The Cls-First scheme surpasses the baseline by 0.3 AP while the Reg-First scheme behinds the baseline by 0.3 AP. The reasons for these results are explained as follows:

- For neural networks, it is widely known that the classification task is much easier than regression task. In Cls-First scheme, the feature passes through more intermediate layers than the baseline to reach the regression output layer. Hence, the network learns more useful features and is more accurate.
- The category information of object provides the prior knowledge of its bounding box as detailed in Section 3.2. In the Cls-First scheme, the category prior feature is fully utilized in the regression head. It is no wonder that the performance of Cls-First scheme is higher than baseline, not to mention Reg-First scheme.

Table3. The performance of two proposed modules. (Backbone: ResNet-50)

| Method | SA-FPN (ALS-Light) | Seq-Head (Cls-First) | AP | $AP_{50}$ | $AP_{75}$ | $AP_S$ | $AP_M$ | $AP_L$ | Params/M |
|---|---|---|---|---|---|---|---|---|---|
| RetinaNet (Anchor-based) | | | 36.1 | 55.1 | 38.4 | 19.4 | 39.6 | 48.8 | 37.7 |
| | ✓ | | 37.0 | 56.4 | 39.7 | 20.3 | 40.9 | **49.4** | 36.5 |
| | | ✓ | 36.4 | 55.6 | 39.1 | 19.8 | 39.6 | 48.2 | 37.7 |
| | ✓ | ✓ | **37.2** | **56.8** | **39.9** | **20.9** | **41.2** | 49.0 | **36.5** |
| FCOS (Anchor-free) | | | 37.0 | 55.8 | 39.7 | 21.0 | 41.0 | 47.7 | 32.0 |
| | ✓ | | 37.4 | **56.5** | 40.0 | 22.0 | 41.4 | 48.2 | 30.8 |
| | | ✓ | 37.6 | 55.8 | 40.6 | 21.7 | 41.3 | **48.5** | 32.0 |
| | ✓ | ✓ | **37.8** | 56.1 | **40.8** | **22.2** | **41.7** | 48.3 | **30.8** |

Table4. The performance of two proposed modules. (Backbone: ResNet-101)

| Method | ALS-Light | Cls-First | AP | $AP_{50}$ | $AP_{75}$ | $AP_S$ | $AP_M$ | $AP_L$ | Params/M |
|---|---|---|---|---|---|---|---|---|---|
| RetinaNet (Anchor-based) | | | 38.9 | 58.2 | 41.7 | 21.5 | 42.9 | 51.4 | 56.6 |
| | ✓ | ✓ | **39.2** | **59.1** | **42.2** | **21.8** | **43.6** | **51.5** | **55.4** |
| FCOS (Anchor-free) | | | 39.3 | 58.2 | 42.3 | 23.0 | 43.7 | 50.6 | 51.0 |
| | ✓ | ✓ | **39.7** | **58.5** | **42.9** | **23.7** | **43.8** | **51.5** | **49.8** |

The Cls-First scheme is selected as the default scheme of sequential heads module. Intuitively, the sequential heads should have much more inference time than parallel heads, it is the reason we conduct experiments to test FPS of both schemes. We do the test experiments on single RTX 2080Ti. The test's input images are resized to $800 \times 1333$, each model is tested for 10 rounds and we take their average as final results. As show in Table 3, both of the two schemes have the same FPS with baseline. We explain it as that the backbone dominates the main inference time of the whole model, and the sequential order of several last layers has only slight effect, which can be ignored.

**Performance of two modules.** We also conduct the combination experiment of two proposed modules. The results are shown in Table 3. The RetinaNet obtains 0.9 AP gain when equipped SA-FPN module alone, and obtain 0.3 AP gain when equipped Seq-Heads module alone. With the help of two modules, the model surpasses baseline by 1.1 AP without extra computational cost. The effectiveness of two modules is proved on anchor-based model, then we carry out experiments on the anchor-free model to prove the universality and robustness of our modules.

### 4.3 Comparison with general networks

In this section, we employ both of the two proposed modules on anchor-based and anchor-free one-stage object detection networks to evaluate the universality and robustness of our modules. We select RetinaNet and FCOS as our anchor-based and anchor-free baseline, respectively. The two networks are both evaluated with ResNet-50 and ResNet-101 backbones, which are pre-trained with ImageNet1k. As shown in Table 3, using ResNet-50 as backbone, the two modules promote AP of anchor-based model RetinaNet by 1.1 points and anchor-free model FCOS by 0.8 AP. We also find out that the SA-FPN module plays a dominant role in improvement of RetinaNet compared with Seq-Head module (0.9 vs 0.3), while the Seq-Head module is dominant power in improvement of FCOS compared with SA-FPN module (0.6 vs 0.4). Table 4 shows that results with ResNet-101 as backbone, RetiaNet and FCOS are improved by 0.3 and 0.4 points AP without extra computational cost, respectively. The improvements on RetinaNet and FCOS prove that the universality of our modules across anchor-based and anchor-free model. Through experiments conducted on networks with ResNet-50 and ResNet-101 as backbones, the robustness is proved. Therefore, our modules can be used in both shallow and deep networks.

## 5 CONCLUSION

In this paper, we present two zero cost modules to improve AP for general objetion detection networks. The SA-FPN module applies scale attention mechanism to original FPN structure. It generates new feature map of each scale by adaptively fusing input feature maps of all scales, according to the attention weights learned from inputs. The sequential head module is used to replace the parallel head module, which is widely used in general object detection networks especially in one-stage domain. The two modules have less or equal parameters compared with their counterparts, but with higher precision. Other application scenarios, such as instance segmentation, about the two modules would be discussed in future work.

## REFERENCES

[1] A. Krizhevsky, I. Sutskever, G. E. Hinton, Imagenet classification with deep convolutional neural networks. Communications of the ACM, Vol.60, No.6, 94-90, 2017.
[2] K. Simonyan, A. Zisserman, Very deep convolutional networks for large-scale image recognition, Computer ence, 2014.

[3] K. He, X. Zhang, S. Ren, J. Sun, Deep residual learning for image recognition, Proceedings of the IEEE conference on computer vision and pattern recognition, 770-778, 2016.

[4] R. Girshick, J. Donahue, T. Darrell and J. Malik, Rich Feature Hierarchies for Accurate Object Detection and Semantic Segmentation, 2014 IEEE Conference on Computer Vision and Pattern Recognition, Columbus, OH, 2014

[5] R. Girshick, Fast R-CNN, 2015 IEEE International Conference on Computer Vision (ICCV), Santiago, 2015

[6] S. Ren, K. He, R. Girshick and J. Sun, Faster R-CNN: Towards Real-Time Object Detection with Region Proposal Networks, in IEEE Transactions on Pattern Analysis and Machine Intelligence, vol.39, no.6, 1137-1149, 2017

[7] K. He, G. Gkioxari, P. Dollár and R. Girshick, Mask R-CNN, 2017 IEEE International Conference on Computer Vision (ICCV), Venice, 2980-2988, 2017

[8] J. Redmon, S. Divvala, R. Girshick and A. Farhadi, You Only Look Once: Unified, Real-Time Object Detection, 2016 IEEE Conference on Computer Vision and Pattern Recognition (CVPR), Las Vegas, NV, 779-788, 2016.

[9] J. Redmon, and A. Farhadi, YOLO9000: better, faster, stronger, Proceedings of the IEEE conference on computer vision and pattern recognition, 2017.

[10] J. Redmon, A. Farhadi, Yolov3: An incremental improvement, arXiv preprint arXiv:1804.02767, 2018.

[11] S. Liu, L. Qi, H. Qin, J. Jia, Path aggregation network for instance segmentation, Proceedings of the IEEE conference on computer vision and pattern recognition, : 8759-8768, 2018.

[12] T. Y. Lin, P. Dollár, R. Girshick, K. He, Feature pyramid networks for object detection, Proceedings of the IEEE conference on computer vision and pattern recognition, 2117-2125, 2017.

[13] C. Zhu, Y. He and M. Savvides, Feature Selective Anchor-Free Module for Single-Shot Object Detection, 2019 IEEE/CVF Conference on Computer Vision and Pattern Recognition (CVPR), Long Beach, CA, USA, 840-849, 2019.

[14] S. Qiao, L. C. Chen, A. Yuille, DetectoRS: Detecting Objects with Recursive Feature Pyramid and Switchable Atrous Convolution, arXiv preprint arXiv:2006.02334, 2020.

[15] T. Y. Lin, M. Maire, S. Belongie, J. Hays. Microsoft coco: Common objects in context, European conference on computer vision, Springer, Cham, 740-755, 2014.

[16] G. Ghiasi, T. Lin and Q. V. Le, NAS-FPN: Learning Scalable Feature Pyramid Architecture for Object Detection, 2019 IEEE/CVF Conference on Computer Vision and Pattern Recognition (CVPR), Long Beach, CA, USA, 7029-7038, 2019.

[17] M. Tan, R. Pang and Q. V. Le, EfficientDet: Scalable and Efficient Object Detection, 2020 IEEE/CVF Conference on Computer Vision and Pattern Recognition (CVPR), Seattle, WA, USA, 10778-10787, 2020.

[18] J. Hu, L. Shen, S. Albanie, G. Sun and E. Wu, Squeeze-and-Excitation Networks, IEEE Transactions on Pattern Analysis and Machine Intelligence, Vol. 42, No. 8, 2011-2023, 2020.

[19] X. Li, W. Wang, X. Hu and J. Yang, Selective Kernel Networks, 2019 IEEE/CVF Conference on Computer Vision and Pattern Recognition (CVPR), Long Beach, CA, USA, 510-519, 2019.

[20] S. Xie, R. Girshick, P. Dollár, Z. Tu and K. He, Aggregated Residual Transformations for Deep Neural Networks, 2017 IEEE Conference on Computer Vision and Pattern Recognition (CVPR), Honolulu, HI, 5987-5995, 2017.

[21] N. Carion, F. Massa, G. Synnaeve, N. Usunier. End-to-End Object Detection with Transformers, arXiv preprint arXiv:2005.12872, 2020.

[22] T. Lin, P. Goyal, R. Girshick, K. He and P. Dollár, Focal Loss for Dense Object Detection, in IEEE Transactions on Pattern Analysis and Machine Intelligence, Vol. 42, No. 2, 318-327, 2020.

[23] Z. Tian, C. Shen, H. Chen and T. He, FCOS: Fully Convolutional One-Stage Object Detection, 2019 IEEE/CVF International Conference on Computer Vision (ICCV), Seoul, Korea (South), 9626-9635, 2019.

[24] Francisco Massa and Ross Girshick. maskrcnn-benchmark: Fast, modular reference implementation of Instance Segmentation and Object Detection algorithms in PyTorch. https://github.com/facebookresearch/maskrcnn-benchmark, 2018.

[25] J. Deng, W. Dong, R. Socher, L. Li, Kai Li and Li Fei-Fei, ImageNet: A large-scale hierarchical image database, 2009 IEEE Conference on Computer Vision and Pattern Recognition, Miami, FL, 248-255, 2009.

[26] A. Paszke, S. Gross, F. Massa, A. Lerer, J. Bradbury, G. Chanan, et al, PyTorch: An Imperative Style, High-Performance Deep Learning Library, 2019.